# Hybrid Genetic Algorithm and Hill Climbing Optimization for the Neural Network

Krutika Sarode
Computer Science
University of Massachusetts Amherst
Amherst MA US
ksarode@umass.edu

Shashidhar Reddy Javaji
Computer Science
University of Massachusetts Amherst
Amherst MA US
sjavaji@umass.edu

## ABSTRACT

In this paper, we propose a hybrid model combining genetic algorithm and hill climbing algorithm for optimizing Convolutional Neural Networks (CNNs) on the CIFAR-100 dataset. The proposed model utilizes a population of chromosomes that represent the hyperparameters of the CNN model. The genetic algorithm is used for selecting and breeding the fittest chromosomes to generate new offspring. The hill climbing algorithm is then applied to the offspring to further optimize their hyperparameters. The mutation operation is introduced to diversify the population and to prevent the algorithm from getting stuck in local optima. The Genetic Algorithm is used for global search and exploration of the search space, while Hill Climbing is used for local optimization of promising solutions. The objective function is the accuracy of the trained neural network on the CIFAR-100 test set. The performance of the hybrid model is evaluated by comparing it with the standard genetic algorithm and hill-climbing algorithm. The experimental results demonstrate that the proposed hybrid model achieves better accuracy with fewer generations compared to the standard algorithms. Therefore, the proposed hybrid model can be a promising approach for optimizing CNN models on large datasets.

## INTRODUCTION

CIFAR-100 is a popular image classification dataset consisting of 60,000 32x32 color images in 100 classes, with 600 images per class. Convolutional Neural Networks (CNNs) are widely used for image classification tasks, including the CIFAR-100 dataset. When using CNNs for CIFAR-100, it is important to design a network architecture that can effectively extract features from the images and classify them into their respective classes.

A genetic algorithm (GA) is a type of optimization algorithm inspired by the process of natural selection and genetics. It is commonly used in the context of neural networks to optimize the weights and biases of a network for a specific task. The GA operates on a population of solutions, each represented as a chromosome. In the context of neural networks, each chromosome typically represents a set of weights and biases for the network. The population is then evolved over a number of generations, with each generation involving selection, crossover, and mutation operations. During the selection phase, individuals in the population with higher fitness scores, which represent how well they perform on the task at hand, are more likely to be selected to breed and create offspring.

The crossover operation involves taking two selected individuals and combining their chromosomes to create new offspring.This is done by randomly selecting a crossover point in the chromosomes and swapping the genes on either side of the point. Finally, the mutation operation involves randomly changing genes in the chromosome to introduce new variations into the population.

This helps to prevent the population from getting stuck in local optima and encourages exploration of the search space. The process of selection, crossover, and mutation is repeated over a number of generations until a stopping criterion is met, such as a maximum number of generations or the attainment of a satisfactory fitness score. In the context of neural networks, the GA can be used to optimize the weights and biases of the network for a specific task. By evolving the population over multiple generations, the GA can explore the search space of potential solutions and find optimal sets of weights and biases that minimize the error on the task. Overall, the GA can be a powerful tool for optimizing neural networks, particularly for complex tasks where traditional optimization techniques may struggle. However, it can also be computationally expensive and may require careful tuning of its parameters to achieve optimal results.

Hill climbing is another optimization algorithm commonly used in the context of neural networks. It belongs to the family of local search algorithms and is used to find the optimal solution in the vicinity of the current solution. In other words, it explores the immediate neighborhood of the current solution and moves in the direction of improvement until it reaches a local maximum. In the context of neural networks, hill climbing can be used to optimize the weights and biases of the network for a specific task. It starts with an initial set of weights and biases and iteratively improves the solution by making small changes to the weights and biases and evaluating the performance of the network on the task. If the new solution is better than the current solution, it becomes the new current solution, and the process continues until a stopping criterion is met. One of the main advantages of hill climbing is its simplicity and efficiency, as it only requires the evaluation of the network performance at each iteration

Hill climbing and genetic algorithms are both search algorithms used in optimization problems, but they differ in their approach and behavior. Here are some of the major differences between the two:

**Approach:** Hill climbing is a local search algorithm that starts with an initial solution and iteratively makes small changes to it to try and improve it. Genetic algorithms, on the other hand, use a population of solutions and evolve them over time by applying genetic operators such as crossover and mutation.

**Exploration vs. Exploitation:** Hill climbing focuses on exploiting the current solution by making small incremental changes to it, whereas genetic algorithms balance between exploration and exploitation by maintaining a diverse population and searching for solutions across the search space.

**Convergence:** Hill climbing can get stuck in local optima and may not be able to find the global optimum if the search space is complex. Genetic algorithms are less likely to get stuck in local optima and can converge to the global optimum over time.

**Scalability:** Hill climbing is suitable for small search spaces and may not be scalable for larger problems. Genetic algorithms can be scaled up to handle larger problems. Performance: Hill climbing can be faster than genetic algorithms in finding a solution in small search spaces. However, genetic algorithms can be more efficient and effective in finding better solutions in complex search space

The idea behind combining global and local optimization algorithms is to leverage the strengths of each approach to improve the overall optimization process. Global optimization algorithms, such as genetic algorithms, are good at exploring the entire search space to find a wide range of potential solutions. However, they can be computationally expensive and may struggle with fine-tuning solutions to reach the optimal point. On the other hand, local optimization algorithms, such as hill-climbing, are good at refining solutions and finding the local optima.

By combining these two approaches, we can start with a global optimization algorithm to explore a wide range of solutions and find promising areas of the search space. Then, we can switch to a local optimization algorithm to fine-tune the solutions and converge toward the optimal point. This approach can help to overcome the limitations of each individual approach and improve the overall optimization process. In the context of neural network optimization, a hybrid approach that combines genetic algorithms with hill-climbing or other local optimization techniques can help to find better sets of weights and biases for the network. The genetic algorithm can be used to explore the search space and identify promising areas, while the local optimization algorithm can be used to fine-tune the solutions and converge toward the optimal point. This can help to improve the performance of the neural network and achieve better results on the given task.

## PROPOSED METHODOLOGY

The architecture is a Convolutional Neural Network (CNN) model for image classification. It consists of several layers that processess the input image in a hierarchical manner, extracting increasingly complex features at each layer.

The first layer is a Conv2D layer with f1 filters and k x k kernel size, followed by another Conv2D layer with the same configuration. Both layers use the same activation function a1. The input shape of the first layer is (32, 32, 3), indicating that the input image has a height and width of 32 pixels, with three color channels (red, green, blue).

After the two Conv2D layers, a MaxPooling2D layer is added with a 2 x 2 pool size, which downsamples the feature maps by taking the maximum value in each 2 x 2 block.Then two more Conv2D layers are added this time with f2 filters and k x k kernel size. Again, both layers use the same activation function a2. Another MaxPooling2D layer is added with the same configuration as before.
The output of the last MaxPooling2D layer is flattened and passed through a Dropout layer with a dropout rate of d1. Then a Dense layer with f3 units and activation function a2 is added, followed by another Dropout layer with a dropout rate of d2. Finally, the output layer is a Dense layer with 100 units and a softmax activation function, indicating that the model is trained to predict one of the 100 possible classes of the CIFAR-100 dataset.

The CNN_model function defines the CNN architecture using the Keras API and compiles it with categorical cross-entropy loss and an optimizer, and trains it on the training set using early stopping based on validation accuracy. The function returns the trained model parameters. The initialization function randomly generates a dictionary of hyperparameters for a CNN. The generate_population function generates a population of n candidate solutions by calling the initialization function n times.

The fitness_evaluation function evaluates the fitness of a given CNN by computing its accuracy on the test set. The selection function chooses parents from the population by picking the top two models with highest accuracy from the population. The crossover function performs a one-point crossover on two parents to produce two children, each inheriting random hyperparameters from both parents.

The mutation function performs a hill climbing step to improve the fitness of the children generated from crossover.The hill climbing step works by evaluating the fitness of the current member and then making small random perturbations to its parameters (i.e., weights and biases) and re-evaluating the fitness. If the fitness improves, the perturbed parameters are kept and the process continues. If the fitness does not improve, the perturbed parameters are discarded and the process starts over with a new set of perturbations. This process continues until no further improvement can be made.

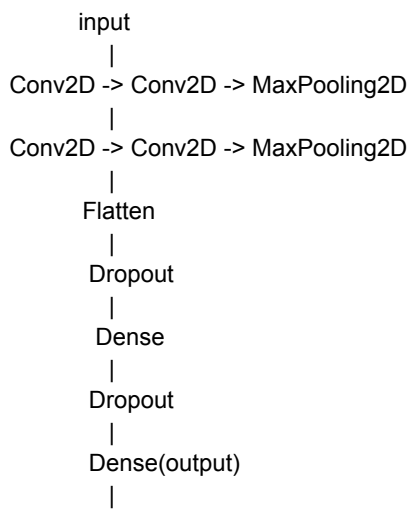

Fig 1: Architecture of CNN model

**Objective:**

The objective of the research is to optimize the hyperparameters of a CNN model to achieve high accuracy on a given task.

**Genetic Algorithm:**

Genetic algorithm is used to evolve a population of candidate solutions (chromosomes) representing different sets of hyperparameters for the CNN model.

The population size is specified by the num_pop variable.

The initialization() function initializes a chromosome with random values for hyperparameters such as filter sizes, activation functions, dropout rates, optimizers, and epochs.

The generate_population() function generates an initial population of chromosomes by calling initialization() multiple times.

The fitness of each chromosome is evaluated using the fitness_evaluation() function, which trains and evaluates a CNN model with the given hyperparameters.

**Selection:**

The selection process determines which chromosomes will be used as parents for creating the next generation.The selection() function selects the two chromosomes with the highest fitness scores from the current population as parents.

**Crossover:**

The crossover operation creates offspring chromosomes by combining the genetic information of the selected parents. The crossover() function performs crossover on selected parents to produce two child chromosomes. It randomly selects hyperparameters from the parents for each child.

**Hill Climbing:**

The hill climbing algorithm is used to perform local search around the best solution found so far. The hill_climbing() function takes the best chromosome from the population and performs hill climbing to explore neighboring solutions by randomly changing a single hyperparameter value at a time.The best neighboring solution with improved fitness is returned as the new best solution.

**Mutation:**

The mutation() function incorporates the hill climbing algorithm to explore new solutions by performing mutations on child chromosomes.For each child, the hill climbing algorithm is applied to find a better solution by perturbing a single hyperparameter at a time.

**Evolutionary Loop:**

The code runs for a specified number of generations. In each generation, the fitness of the population is evaluated, and the best chromosome's fitness is checked against the threshold for terminating the algorithm. If the threshold is reached, the algorithm terminates and outputs the best solution. Otherwise, the algorithm continues to the next generation.

The two worst-performing chromosomes are removed from the population to maintain a constant population size. The offspring chromosomes (children) generated through crossover and mutation are added to the population.

**Output and Termination:**

The code prints the information about the best solution (chromosome) found in each generation, including the hyperparameters and the corresponding accuracy achieved. If the desired accuracy threshold is reached, the algorithm terminates and outputs the best solution.

## Data:

CIFAR-100 is a widely used dataset for image classification tasks in computer vision research. It contains 60,000 32x32 color images in 100 classes, with 600 images per class. The dataset is split into 50,000 training images and 10,000 testing images. Each image in the CIFAR-100 dataset belongs to one of 100 fine-grained classes that are grouped into 20 coarse-grained classes. The fine-grained classes represent specific objects or organisms, while the coarse-grained classes represent general categories such as mammals, birds, fish, flowers, and trees. The dataset is designed to test the ability of machine learning algorithms to classify images at different levels of granularity.

CIFAR-100 is a challenging dataset for image classification because the images are small and low-resolution, and the classes are highly diverse and fine-grained. Many state-of-the-art deep learning models have been trained on the CIFAR-100 dataset, and it is commonly used as a benchmark for evaluating the performance of new models.

To use the CIFAR-100 dataset in research, one can download it from the official website or use a pre-processed version from a library such as PyTorch or TensorFlow. The dataset can be split into training, validation, and test sets, and pre-processing steps such as normalization and data augmentation can be applied to improve the performance of machine learning models. In research papers, the CIFAR-100 dataset is often used to evaluate the performance of new machine learning models, compare different models or architectures, or analyze the behavior of models under different conditions. Researchers may also use the dataset to perform ablation studies, where different components of a model or algorithm are systematically removed to understand their contribution to the overall performance.

## Baselines:

Our research project aims to compare the performance of two distinct convolutional neural networks (CNN) models trained on the CIFAR-100 dataset. We consider a baseline CNN model without any optimization techniques, as well as two variants incorporating heuristic search algorithms: one utilizing hill climbing and the other employing a genetic algorithm. Our goal is to evaluate how these models differ in their accuracy on the CIFAR-100 dataset, with the expectation that the hybrid approach leveraging both hill climbing and genetic algorithms will yield the highest level of performance. Our research findings will contribute to the ongoing efforts to optimize the training of CNN models for image classification tasks.

## Experiments and Results:

The model is experimented with the architecture as proposed, though the results are better than the usual genetic algorithm, the compute resource required is very high. there is a decision one has to make with compute resources against the required accuracy. The results are as follows.

|  | Genetic Algorithm | Hybrid Model |
| --- | --- | --- |
| After Generation 1 | 0.393 | 0.470 |
| After Generation 2 | 0.466 | 0.481 |
| After Generation 3 | 0.459 | 0.473 |
| Total Runtime | 1 hr 20 min | 4 hr 30 min |

Table: Performance of Models

The results show that the performance has increased, though the difference is not very significant, we ran it for only 3 generations and 5 populations,

The genetic algorithm achieved varying levels of accuracy throughout the iterations. Indicating that different sets of parameters and configurations led to different levels of performance.

The hill climbing algorithm, when applied to the best individuals obtained from the genetic algorithm, improved the fitness or accuracy in most cases. The improvements ranged from slight enhancements to more significant gains.

The hybrid model, combining the genetic algorithm with hill climbing, showed promising results compared to the genetic algorithm alone. The accuracy values obtained by the hybrid model were generally higher than those of the genetic algorithm.

The hill climbing algorithm applied to the best individual from the hybrid model also showed some improvements, although

**Genetic Algorithm:**

Under the "Genetic Algorithm" column, the accuracy values achieved by the GA are provided. After the first generation, the GA achieved an accuracy of 0.393. In the second generation, the accuracy improved to 0.466, and in the third generation, it slightly decreased to 0.459.

**Hybrid Model:**

Under the "Hybrid Model" column, the accuracy values achieved by the hybrid model (GA combined with another technique) are presented. After the first generation, the hybrid model achieved an accuracy of 0.470, which is higher than the GA's accuracy of 0.393. In the second generation, the hybrid model further improved its accuracy to 0.481. However, in the third generation, the accuracy decreased slightly to 0.473.

**Total Runtime:**

The "Total Runtime" row provides information about the time it took to run the algorithms. The genetic algorithm took a total of 1 hour and 20 minutes to complete, while the hybrid model required more time, totaling 4 hours and 30 minutes.

In summary, the table demonstrates the performance of the genetic algorithm and the hybrid model over multiple generations. The hybrid model shows higher accuracy than the genetic algorithm in most cases, indicating that the incorporation of the additional technique (e.g., hill climbing) has improved the optimization process. However, the accuracy fluctuates across generations for both approaches, suggesting that further analysis and refinement may be required to enhance the overall performance. Additionally, the hybrid model requires more computational time compared to the genetic algorithm, likely due to the additional steps involved in the hybridization process.

## Conclusion and Future Work:

The conclusion that can be made from this experiment is that we can use more hybrid models which can make best of localized and globalized, one thing that has to be taken care of and is to be kept in mind is that these hybrid models take more time to run than the usual models, there can be many others models that can be combined to give better results, for this model at least the result was just good enough and not great improvement, but with more compute resources we hope to see how the model results change.The future works include trying more different algorithms instead of hill climbing and trying some other localized algorithms.